\documentclass[onefignum,onetabnum]{siamonline250211}
\usepackage[T1]{fontenc}
\usepackage{lmodern,amsfonts,amssymb,booktabs,microtype}
\usepackage{enumitem}
\setlist[enumerate]{label=\arabic*.,leftmargin=2.2em,labelsep=0.6em,itemsep=0pt,topsep=0.3\baselineskip}
\hypersetup{hidelinks}
\newsiamremark{remark}{Remark}
\newcommand{\Z}{\mathbb Z}
\newcommand{\Q}{\mathbb Q}
\newcommand{\R}{\mathbb R}
\newcommand{\D}{\mathbb Z[1/2]}
\newcommand{\F}{\mathbb F}
\newcommand{\La}{\mathcal L}
\DeclareMathOperator{\conv}{conv}
\DeclareMathOperator{\sgn}{sgn}

\DeclareMathOperator{\Tot}{Tot}
\newcommand{\PaperTitle}{Integral Multipliers and Arithmetic Obstructions for Two-Hidden-Layer ReLU Maxima}
\newcommand{\AuthorName}{Pengcheng Cheng}
\newcommand{\AuthorAffiliation}{School of Mathematics, Jilin University}
\newcommand{\AuthorAddress}{Changchun 130012, China}
\newcommand{\AuthorEmail}{3192618014@qq.com}
\newcommand{\AuthorList}{\AuthorName\thanks{\AuthorAffiliation,
\AuthorAddress\ (\email{\AuthorEmail}).}}
\newcommand{\RunningAuthors}{Pengcheng Cheng}
\newcommand{\PDFAuthors}{Pengcheng Cheng}

\title{\PaperTitle}
\author{\AuthorList}
\headers{Arithmetic of shallow rectifier networks}{\RunningAuthors}
\ifpdf
\hypersetup{pdftitle={\PaperTitle},pdfauthor={\PDFAuthors},pdfkeywords={ReLU networks, integer weights, lattice polytopes, cubical chains}}
\fi
\begin{document}
\maketitle
\begin{abstract}
We study exact representations of maximum functions by ReLU networks
with two hidden layers and integer parameters. The least positive output
multiplier is two for the six-input maximum, and either two or six for
the seven-input maximum. We obtain these bounds from integral cubical
fillings and a coefficient-preserving passage from polytope identities
to networks. For the coordinate-facet complex of the six-simplex, we
compute a coning index of six over the integers and three after inverting
two, allowing arbitrary integral directions within each facet. A class
in the homology of the unrestricted kernel complex characterizes the
remaining seven-input problem; its vanishing is equivalent to an integral
filling of a fixed target. We also construct a finite character that
separates rational
from three-local representability in a dictionary containing two known
constructions and all pairs of edge-disjoint three-edge forests. This
obstruction permits nonsymmetric coefficients and is confined to the
specified dictionary. Control of lower-dimensional remainders yields a
sufficient criterion in the top-dimensional quotient. Exact certificates
and verification programs accompany the finite calculations.
\end{abstract}
\section{Introduction}

The depth required for exact representation by a ReLU network depends
on the arithmetic of its parameters. Over the reals or the rationals,
two hidden layers suffice for every continuous piecewise-linear function
with data in the same field
\cite[Theorem 1]{brunck}. Over the integers, lattice-polytope obstructions
give a different depth hierarchy \cite{hhl,ahm}. We ask how much of this
obstruction can be removed by an integer output multiplier. For the
maximum function, what is the least positive multiplier that permits
a representation with two hidden layers and integer parameters?

The natural setting for this question is the additive geometry of lattice
polytopes. A maximum of linear forms is a polytope support function, and
a ReLU gate corresponds to a convex-hull operation. Passing to support
functions modulo lower-dimensional polytopes isolates the top-dimensional
geometry; retaining integer coefficients records its arithmetic. This
approach to exact expressivity builds on
\cite{hertrich2023depth,hhl,brunck}; polyhedral decompositions of piecewise-linear
functions also distinguish fixed representation families from
unrestricted representations \cite{brandenburg2026decomposition}.
We use integral coning and explicit cubical fillings to determine the
six-input multiplier and reduce the seven-input problem to one prime.

\subsection{Model and principal results}

Write $\sigma(t)=\max(0,t)$, applied coordinatewise, and set
\[
 M_n(x)=\max(x_1,\ldots,x_n),\qquad
 f_d(y)=\max(0,y_1,\ldots,y_d).
\]
A network with two hidden layers and zero biases has the form
\begin{equation}\label{eq:model}
 F(x)=c^{\mathsf T}\sigma(B\sigma(Ax)),
\end{equation}
with arbitrary finite hidden widths. All three arrays, including the
output weights, belong to the specified coefficient ring. The output
linear map is not counted as a hidden layer. Linear terms are carried
through the hidden layers by $t=\sigma(t)-\sigma(-t)$, so no skip
connections are needed. Parallel sums, subtraction, and integer output
scaling preserve the depth.

The integral changes of variables
\begin{equation}\label{eq:translation}
\begin{aligned}
 f_d(y)&=M_{d+1}(y_1,\ldots,y_d,0),\\
 M_{d+1}(x)&=x_{d+1}+f_d(x_1-x_{d+1},\ldots,x_d-x_{d+1})
\end{aligned}
\end{equation}
transfer multipliers between the targets. Biases do not change the
existence question for a positively homogeneous target: evaluate a
representing network at $tx$, divide by $t$, and let $t$ tend to infinity.
Positive homogeneity of ReLU removes the biases and leaves the weight
matrices unchanged.

The integers $m$ for which $mf_d$ admits an integer representation
\eqref{eq:model} form a nonzero ideal. Nontriviality follows from rational
representability and positive denominator clearing, layer by layer. We
write $E_d$ for its positive generator. This is the arithmetic depth
exponent of Tallavarjula--Rastegar \cite[Section 9]{tr}; their
Proposition 17 gives $E_{d-1}\mid E_d\mid d!$. The index records the dimension:
$E_5$ concerns six inputs and $E_6$ concerns seven inputs.

\begin{theorem}[Integer multipliers]\label{thm:main}
There are integer-weight, zero-bias networks with two hidden layers
representing $2M_6$ and $6M_7$. The least positive integer multipliers
satisfy
\begin{equation}\label{eq:main-exponents}
 E_5=2,\qquad 2\mid E_6\mid6.
\end{equation}
In particular, $E_6\in\{2,6\}$. Both constructions are finite and
algorithmic.
\end{theorem}

The construction uses integral cubical chains whose cones are support
functions of integer two-layer gates. Passing from the rational polytope
complex to an identity of support functions leaves a lower-dimensional
remainder. Proper-face completion retains the integer coefficients of
this remainder; an integral normal form gives the converse passage.
Together they yield
\[
 E_d=\operatorname{lcm}(E_{d-1},G_d),
\]
where $G_d\Z$ is the ideal of multiples of the simplex top class realized
by integral paramid chains with unrestricted lattice directions
(Proposition~\ref{prop:multiplier-lcm}). The rank-reduction operators
underlying completion are due to Zimmermann and Myrisiotis
\cite{zimmermann,witnessing}; we state and prove the coefficient-tracking
form needed here. The converse tracks integrality through the
subdivisions of Haase--Hertrich--Loho \cite{hhl} and the paramid identity
of Brunck et al.\ \cite{brunck}.

We next determine the coning image of the coordinate-facet complex.
Let $A$ consist of cubical generators supported in coordinate
facets of the six-simplex, let $b$ denote its oriented boundary, and let
$\rho$ be coning. Theorem~\ref{thm:index} proves
\[
 \rho Z_4(A;\Z)=6\Z b,\qquad
 \rho Z_4(A;\D)=3\D b.
\]
Arbitrary integral directions are allowed within each coordinate facet.
The two factors arise from differentials in a \v{C}ech spectral sequence;
the order-three differential survives the indeterminacy of the preceding
one. An explicit augmentation repair with three additional directions
then converts the coordinate cycle into an integral filling for $6M_7$.
Theorem~\ref{thm:three-directions-minimal} of the supplement proves that
three directions are necessary for this particular repair.

For unrestricted directions, the obstruction is a class annihilated
by three in a global kernel homology group
(Theorem~\ref{thm:global-theta}). Equivalently, it is an integral filling
problem for a fixed, explicitly recorded target
(Proposition~\ref{next:integral-target}). Deciding whether this class
vanishes would distinguish $E_6=2$ from $E_6=6$; the coordinate-facet
calculation alone does not decide the question.

Finally, Theorem~\ref{thm:finite-separation} separates rational from
$\Z_{(3)}$-representability in a finite dictionary containing two existing
seven-input constructions and a complete family of pairs of
edge-disjoint three-edge forests. The separating character annihilates
each generator, so no symmetry assumption on the coefficients is needed.
An omitted lower-dimensional gate has nonzero character value.
Proposition~\ref{prop:top-criterion} describes the effect of adjoining
all lower-dimensional remainders.

\subsection{Comparison with prior constructions}

The identities of Schroeder and Myrisiotis give integer representations
of $6M_6$ \cite{schroeder,hexagon}; Theorem~\ref{thm:main} gives the optimal
multiplier in this dimension. For seven inputs, the general factorial
bound $E_d\mid d!$ gives $720M_7$ \cite[Proposition 17]{tr}, compared with
the integral construction of $6M_7$ here. The factorial bound provides a
uniform comparison, independent of any particular rational certificate.

Rue\ss{} et al.\ construct rational representations through twelve
inputs \cite{ruess}, Wang--Basu through eight inputs \cite{wangbasu}, and
Zimmermann gives a general Caterpillar Net \cite{zimmermann}. These
constructions, together with field-valued universality \cite{brunck},
settle existence over a field. Their arithmetic requires a separate
analysis: denominator clearing gives an admissible multiplier, whereas
its optimality is a question about the entire integer module. The finite
separation theorem addresses this distinction for a specified dictionary.
The localization criterion of
Tallavarjula--Rastegar \cite[Theorem 21]{tr} gives the following consequence.

\begin{corollary}[Localization]\label{cor:localization}
Let $R=\Z[\mathcal P^{-1}]\subseteq\Q$, with $\mathcal P$ a set of
primes. The function $M_7$ admits a two-hidden-layer representation over
$R$ exactly when every prime divisor of $E_6$ belongs to $\mathcal P$.
It is therefore impossible if $2\notin\mathcal P$, possible if
$2,3\in\mathcal P$, and equivalent to dyadic representability if
$2\in\mathcal P$ but $3\notin\mathcal P$.
\end{corollary}

\begin{proof}
Clearing the finitely many parameter denominators by positive
homogeneity gives an integer network for $mf_6$, with all prime divisors
of $m$ in $\mathcal P$. Hence $E_6\mid m$. Conversely, divide the final
output weights of an integer network for $E_6f_6$ by $E_6$ whenever it is
a unit of $R$. Apply \eqref{eq:main-exponents}.
\end{proof}

\subsection{Organization and verification}

Sections~\ref{sec:prelim}--\ref{sec:completion} establish the integral
geometric framework. Section~\ref{sec:coordinate} computes the
coordinate-facet index, and Section~\ref{sec:positive} proves
Theorem~\ref{thm:main}. Section~\ref{sec:relative} identifies the exact
global obstruction; Section~\ref{sec:finite-separation} treats the finite
dictionary and top-class search criterion. The proofs combine geometric
arguments with finite checks of integer identities and finite-field
pairings. Section~\ref{app:certificates} of the supplement specifies the
machine-readable witnesses and independent verification programs for
these checks, all of which use exact arithmetic.

The supplement also contains extensions to split hyperplanes, the
minimality proof for the three-direction repair, consequences for dyadic
depth, and integer evaluation sets that preserve exact row modules.
It also gives an independently expanded weight array for $24M_7$, valid
on the whole input space (Theorem~\ref{thm:literal}). The construction of
$6M_7$ instead proceeds through an explicit integral filling and
constructive completion; its full weight arrays are not expanded.
\section{Support functions and integral cubical chains}\label{sec:prelim}

\subsection{The rational target and the integral source}

For a polytope $P\subseteq\R^d$, write
$h_P(y)=\max_{v\in P}\langle v,y\rangle$. Denote by $\La_k$ the real
linear span of support functions of polytopes of affine dimension at
most $k$. Thus $\La_0$ is the space of linear functions. We use
the rational oriented-polytope complex $P_\bullet$ of
\cite[Sections 2--3]{brunck}. Its component in an oriented $k$-space
consists of formal Minkowski differences modulo polytopes in
proper direction spaces. Reversing the orientation changes the sign.
The boundary is the sum of the facet classes, oriented by the
outward-normal-first convention. Translations do not affect the
classes in the degrees considered below.

We use the following two properties of this complex:
\begin{equation}\label{eq:geometric-input}
 d\rho=\rho d\quad(q\geq2),\qquad
 d:P_k(H)\longrightarrow P_{k-1}(H)
 \text{ is injective if }\dim H=k\geq2.
\end{equation}
These are Lemmas 9 and 3 of \cite{brunck}. Equality of
top-dimensional classes determines a remainder in $\La_{d-1}$,
but does not imply an integral decomposition of that remainder.
The required integral completion is proved in
Section~\ref{sec:completion}.

Let $\Gamma$ be a lattice, and let $T$ be a set of integral directions
containing a lattice basis. All chains have finite support.
For $R=\Z$ or a localization of $\Z$, the cubical group
$C_q(\Gamma,T;R)$ is free on generators
$(\gamma;t_1,\ldots,t_q)$, alternating in the directions, with
\begin{align}\label{eq:cubical-boundary}
 d(\gamma;t_1,\ldots,t_q)
 =\sum_{j=1}^q(-1)^{j-1}\bigl[&
 (\gamma+t_j;t_1,\ldots,\widehat t_j,\ldots,t_q)\notag\\[-2pt]
 &-(\gamma;t_1,\ldots,\widehat t_j,\ldots,t_q)\bigr].
\end{align}
The augmentation $\epsilon$ sends a generator to its exterior
direction tuple, forgetting the base point. Write
\[
 W_T=\ker\bigl(R^{(T)}\longrightarrow\Gamma\otimes R\bigr),
 \qquad e_t\longmapsto t.
\]

\begin{lemma}[Integral homology coordinates]\label{lem:koszul}
Augmentation induces the isomorphism
\[
 H_q(C_\bullet(\Gamma,T;R))\cong\Lambda^qW_T.
\]
In particular, every cycle with zero augmentation is a boundary
over $R$. For the root-containing coordinate subcomplexes used
below, both the identification and the fillings can be chosen within
the same coordinate set $S$.
\end{lemma}

\begin{proof}
The complex is the Koszul complex over $R[\Gamma]$ on the elements
$g_t-1$, where $g_t$ denotes translation by $t$. Choose a lattice
basis from $T$. For each remaining direction, a finite lattice path
expresses $g_t-1$ as an $R[\Gamma]$-linear combination of the basis
differences. Subtract the corresponding path chain from the
degree-one generator. This invertible change of generators identifies
the complex with the tensor product of the basis Koszul complex and
an exterior algebra on closed relation generators. Successive
contraction of the one-dimensional lattice complexes shows that
the basis Koszul complex has homology $R$ in degree zero and
vanishing homology in positive degrees. Augmentation sends the
remaining generators to a basis of $W_T$.
For a coordinate set, choose a root-star lattice basis in that set;
all paths and contractions then remain within the same coordinate
subcomplex.
\end{proof}

The coning map sends a cubical generator to the oriented polytope
\begin{equation}\label{eq:paramid}
 P_{\gamma,I}=\conv\left(0,\gamma+\sum_{i\in I}[0,t_i]\right),
 \qquad\text{orientation }\bigwedge_{i\in I}t_i\wedge\gamma.
\end{equation}
Degenerate cones vanish in the indicated graded component. Following
\cite{brunck}, we call \eqref{eq:paramid} a \emph{paramid}. Its
support function is
\begin{equation}\label{eq:paramid-support}
 h_{P_{\gamma,I}}(y)=
 \sigma\left(\langle\gamma,y\rangle+
                 \sum_{i\in I}\sigma(\langle t_i,y\rangle)\right).
\end{equation}
This function has an integer two-hidden-layer realization whenever
$\gamma,t_i$ are integral, with signed ReLU pairs carrying the inner
linear term. The coefficient of a full-dimensional coning term in
the corresponding support-function sum is its chain coefficient
multiplied by
$\sgn\det(t_{i_1},\ldots,t_{i_{d-1}},\gamma)$.
The determinant magnitude contributes no multiplicity.

\subsection{Simplex coordinates and reflection}

For $V=\{0,\ldots,d\}$, put
\[
 \Gamma_V=\{a\in\Z^{d+1}:\textstyle\sum_i a_i=0\},\qquad
 r_{ij}=e_j-e_i\quad(i<j).
\]
Deleting coordinate zero identifies $\Gamma_V$ with $\Z^d$.
Under this identification, the simplex $\Delta_V$ has support $f_d$. For
$S=\{s_0<\cdots<s_k\}$, orient $\Delta_S$ by
$r_{s_0s_1}\wedge\cdots\wedge r_{s_0s_k}$, translating its
vertices into $\Gamma_S\otimes\Q$ as necessary.

\begin{lemma}[Reflection]\label{lem:reflection}
Define
\[
 J_q(\gamma;t_1,\ldots,t_q)
 =(-1)^q(-\gamma-t_1-\cdots-t_q;t_1,\ldots,t_q).
\]
Then $J$ is a chain map and $\epsilon J_q=(-1)^q\epsilon$.
In degree four, $\rho J_4=-\rho$.
\end{lemma}

\begin{proof}
The chain-map assertion follows directly from the boundary formula.
Geometrically, this is the oriented cubical map induced by
$\gamma\mapsto-\gamma$. To compute its coning image, we first show
that, in the top quotient of an oriented $k$-space with fixed
orientation,
\begin{equation}\label{eq:antipodal}
 [-P]=(-1)^{k+1}[P].
\end{equation}
For $k=1$, reflection sends a segment to a translate of itself.
For the induction step, reflection negates the outward normal and
hence the induced facet orientation. The $(k-1)$-dimensional
identity shows that the two sides of \eqref{eq:antipodal} have the
same boundary. Top-boundary injectivity proves the identity for
$k\geq2$.
For a degree-four cube, reflection negates both its base and its
coning orientation, while \eqref{eq:antipodal} leaves the
five-dimensional support class unchanged. Hence
$\rho J_4=-\rho$, with the same conclusion for degenerate terms.
\end{proof}

\section{Coefficient-preserving proper-face completion}\label{sec:completion}

Proper-face completion with integer coefficients follows by retaining
each original coefficient throughout the rank-reduction procedure of
Zimmermann \cite[proof of Theorem 7, Step 0.1]{zimmermann} and
Myrisiotis \cite[Appendix C.4, pp. 92--94]{witnessing}. We give a
self-contained proof, including a finite second pass that uses only
the original polytopes. The face operators and the search for integral
directions follow these earlier methods. The subsequent paramid
reduction also preserves integer coefficients.

\begin{lemma}[Integral completion]\label{lem:completion}
Let $P_1,\ldots,P_N$ be full-dimensional lattice polytopes in $\R^d$,
and let $a_i\in\Z$. If $F=\sum_i a_i h_{P_i}$ belongs to
$\La_{d-1}$, then
\begin{equation}\label{eq:completion}
 F=\ell+\sum_i a_i\sum_{\varnothing\ne Q\lneq P_i}b_{i,Q}h_Q,
 \qquad b_{i,Q}\in\Z,
\end{equation}
where $\ell$ is an integer linear form and $Q$ ranges over proper
faces. These coefficients can be computed by a finite deterministic
algorithm from the $P_i$ and $a_i$ alone. The faces of each polytope
retain its original coefficient $a_i$.
\end{lemma}

\begin{proof}
For a nonzero vector $u$, let $P^u$ be the face exposed by $u$, and
define $T_u h_P=h_{P^u}$. The formula
\begin{equation}\label{eq:face-limit}
 T_uF(y)=\lim_{t\to\infty}\bigl(F(tu+y)-tF(u)\bigr)
\end{equation}
shows that $T_u$ is a linear operator on support-function
combinations, independent of the chosen presentation.

Express $F$ as a finite real linear combination of lower-dimensional
support functions, and close this family of polytopes under nonempty
faces. For each distinct
$k$-dimensional direction space $W$ in this family, choose
$u_W\in W^\perp\setminus\{0\}$ outside the orthogonal complement
of every other $k$-dimensional space in the family. Such a choice
exists because a finite union of proper subspaces cannot cover
$W^\perp$.
Set
\[
 A_k=I-\sum_{\dim W=k}T_{u_W},\qquad C=A_1A_2\cdots A_{d-1},
\]
with the rightmost factor acting first. On a polytope with direction
space $W$ of dimension $k$, the identity term cancels the
$T_{u_W}$ term, leaving only proper faces. Terms of smaller
dimension remain of smaller dimension. Applying
$A_{d-1},\ldots,A_1$ successively therefore makes $CF$ linear.

Now apply this operator to the original full-dimensional
presentation. Every nonidentity word in the $T_u$ selects a proper
face at its first step, since the first direction is nonzero;
subsequent steps select faces contained in that face. Expanding
$CF-F$ gives an integer combination of proper faces of each $P_i$,
retaining the coefficient $a_i$. All these faces have integer
vertices, so $CF(e_j)\in\Z$ for every coordinate vector $e_j$.
The linear function $CF$ is therefore integral. Rearrangement yields
\eqref{eq:completion}.

A second pass removes the dependence on the auxiliary decomposition.
The preceding argument places $F$ in the integer span of the proper
faces of the original $P_i$ and integer linear functions. Define
$A'_k$ and $C'$ by the same construction, using the direction spaces
of these finitely many faces. The rank reduction again makes
$C'F$ linear. Compute $C'F$ from the original presentation, expand
$(I-C')F$, and determine the integral linear remainder from
$C'F(e_j)$.

All direction spaces in the second pass are rational. To choose
an admissible integral $u_W$, take an integer spanning basis
$z_0,\ldots,z_{c-1}$ of $W^\perp$ and try
$\sum_{j=0}^{c-1}t^jz_j$ for integers $t\geq0$. Each forbidden
subspace excludes only roots of a nonzero polynomial, so the search
terminates. The selected directions enter only the face operators;
they contribute no network weights. The operator $C'$ need not fix
linear functions, since each $A'_k$ multiplies them by an integer
scalar. Only preservation of linearity is required.
\end{proof}

\begin{corollary}[From simplex classes to networks]\label{cor:bridge}
Suppose an integral $(d-1)$-chain $u$ satisfies
$\rho u=m\Delta_V$, where $m>0$. Suppose also that $m$ times the
maximum of any at most $d$ integer linear forms has an integer
two-hidden-layer representation. Then $mM_{d+1}$ has such a
representation.
\end{corollary}

\begin{proof}
Let $g_u$ be the integer support-function sum of the full-dimensional
coning terms of $u$, with the determinant signs specified after
\eqref{eq:paramid-support}. Equality in the rational polytope complex
gives $mf_d-g_u\in\La_{d-1}$, to which we apply
Lemma~\ref{lem:completion}.
Every proper face of a full-dimensional paramid is a point, a
translated parallelepiped face, or the cone from its apex over a
proper base face. Its support therefore has an integer realization
of depth at most two, using \eqref{eq:paramid-support} and signed
linear pairs.
The proper simplex faces have at most $d$ vertices, and their
coefficients retain the factor $m$. The assumed representations
therefore realize every simplex term. Summing the resulting networks
and using \eqref{eq:translation} completes the proof.
\end{proof}

\subsection{An integral normal form for the top class}
\label{subsec:integral-paramid-normal-form}

The paramid universality argument of \cite[Section 5.1]{brunck}
can be carried out with lattice vertices and integer coefficients.
The top support-function quotient remains the quotient by the real
space $\La_{d-1}$; integrality concerns the coefficients and vertices
of the resulting full-dimensional presentation.
Let
\[
 C_q^{\mathrm{all}}(\Z^d;R)
 =C_q(\Z^d,\Z^d\setminus\{0\};R).
\]
This is the filtered union over finite direction sets containing a
lattice basis, and every chain has finite support.

\begin{lemma}[Integral paramid normal form]
\label{lem:integral-paramid-normal-form}
Let $d\ge2$.  If a positively homogeneous function $F$ on $\R^d$
has an integer two-hidden-layer ReLU representation, then its
top support-function class is $\rho u$ for some
$u\in C_{d-1}^{\mathrm{all}}(\Z^d;\Z)$.
If every nonzero first-layer row is an integer multiple, of either
sign, of a vector in a prescribed set $T$ of primitive lattice
directions, then $u$ may be chosen in $C_{d-1}(\Z^d,T;\Z)$.
In particular, an integer representation of $mf_d$ implies
\[
 m\Delta_d\in\rho C_{d-1}^{\mathrm{all}}(\Z^d;\Z).
\]
\end{lemma}

\begin{proof}
Homogeneity allows us to remove any biases. Write a second-layer
preactivation as $h_{Z_+}-h_{Z_-}$, where $Z_+$ and $Z_-$ are
lattice zonotopes: a term $b\sigma(\langle v,x\rangle)$ contributes
the lattice segment $[0,bv]$ to $Z_+$ if $b\ge0$, and
$[0,-bv]$ to $Z_-$ otherwise. Thus
\[
 \sigma(h_{Z_+}-h_{Z_-})
 =h_{\operatorname{conv}(Z_+\cup Z_-)}-h_{Z_-}.
\]
The second term is a sum of segment supports, hence lies in
$\La_1\subseteq\La_{d-1}$. It suffices to express the top class of
$P=\operatorname{conv}(Z_+\cup Z_-)$ as an integral sum of paramid
classes, since the output coefficients are integers.
The case $\dim P<d$ contributes zero, so assume $\dim P=d$.

The subdivision in \cite[Proposition 10]{hhl} expresses a
full-dimensional $P$, modulo lower-dimensional terms, as the sum
of full-dimensional joins $\operatorname{conv}(A\cup B)$ of faces
of $Z_+$ and $Z_-$. Each nonempty $A$ and $B$ is a lattice
zonotope. A full-dimensional cell with one empty side is itself
a zonotope, so its top support class vanishes. In the remaining
cells, the affine hulls of $A$ and $B$ are disjoint and their direction
spaces are transverse, with
$\dim A+\dim B=d-1$.

To subdivide a lattice zonotope into lattice parallelepipeds,
apply the Minkowski-sum subdivision of \cite[Proposition 9]{hhl}
repeatedly to its segment summands. The full-dimensional cells
are sums of independent segments and segment endpoints, and thus
have lattice vertices. Subdivide $A$ and $B$ in this way inside
their join. The joins of their cells cover the original join with
disjoint relative interiors, and intersections of distinct
full-dimensional cells have lower dimension. The valuation
relation therefore assigns coefficient $+1$ to each full-dimensional
join of parallelepipeds. Generic lifts affect only the auxiliary
heights; every projected vertex remains a sum of original vertices.
These subdivisions require no denominator clearing.

Consider such a join with $A=a_0+A_0$, where $A_0$ is a lattice
parallelepiped and $a_0\in\Z^d$. The reflection identity
\cite[Lemma 14]{brunck}, applied to the Minkowski decomposition
$A=\{a_0\}+A_0$, gives
\[
 [\operatorname{conv}(A\cup B)]
 =(-1)^{\dim A}
   [\operatorname{conv}(\{a_0\}\cup(B-A_0))]
 =(-1)^{\dim A}
   [\operatorname{conv}(\{0\}\cup(B-A))].
\]
The second equality is translation by the lattice vector $-a_0$,
which changes the support function by a linear form. Here $B-A$
denotes the Minkowski sum $B+(-A)$. Transversality of the direction
spaces of $A$ and $B$ makes $B-A$ a lattice
$(d-1)$-parallelepiped. Disjointness of the affine hulls gives
$0\notin\operatorname{aff}(B-A)$, so the last polytope is a
full-dimensional lattice paramid. Reflection introduces only the
coefficient $(-1)^{\dim A}$.

Write each base as
$\gamma+\sum_{j=1}^{d-1}[0,t_j]$, with
$\gamma,t_j\in\Z^d$, and orient the corresponding cube so that
its coning orientation agrees with the ambient orientation.
In coefficients, this amounts to multiplication by
$\operatorname{sgn}\det(t_1,\ldots,t_{d-1},\gamma)$.
Only the determinant sign enters this conversion. The sum of these
cubes is the required finite chain $u$.

For prescribed directions, every zonotope segment is an integer
multiple of an element of $T$. The Minkowski subdivisions select
independent segments from the original summands, and the base $B-A$
of each reflected join has only these edge directions. Reverse
negative segments, translating their base points by lattice vectors, and
subdivide each parallelepiped along its integer edge lengths into
parallelepipeds with unit generators in $T$. Coning this subdivision
from the apex preserves the top class and has integer coefficients,
since the base hyperplane does not contain the apex. This identity
holds in the top quotient: it does not identify a long-edge cubical
generator with a unit-edge chain before applying $\rho$.
The direction-preserving normal form requires no lattice-basis
hypothesis on $T$.
\end{proof}

Define the top-class multiplier ideal
\[
 \mathcal I_d^{\mathrm{par}}=
 \{m\in\Z:m\Delta_d\in
       \rho C_{d-1}^{\mathrm{all}}(\Z^d;\Z)\}
 =G_d\Z,
\]
where $G_d>0$. To see that the ideal is nonzero, clear positive
denominators in a rational representation to obtain an integer
network for a positive multiple of $f_d$, and apply
Lemma~\ref{lem:integral-paramid-normal-form}.

\begin{proposition}[Top-class and face multipliers]
\label{prop:multiplier-lcm}
For $d\ge2$, the arithmetic exponents satisfy
\begin{equation}\label{eq:multiplier-lcm}
 E_d=\operatorname{lcm}(E_{d-1},G_d).
\end{equation}
Consequently, with $E_1=1$,
$E_d=\operatorname{lcm}(G_2,\ldots,G_d)$.
\end{proposition}

\begin{proof}
If $mf_d$ has an integer two-hidden-layer representation, then
Lemma~\ref{lem:integral-paramid-normal-form} gives $G_d\mid m$.
Restricting to $x_d=0$ gives an integer representation of
$mf_{d-1}$, so $E_{d-1}\mid m$ as well.
Conversely, suppose $m>0$ is divisible by $G_d$ and $E_{d-1}$.
Choose an integral chain $u$ with $\rho u=m\Delta_d$.
Every proper simplex face has at most $d$ vertices, so $m$ times
its support has an integer two-hidden-layer representation by
integral input substitution in a representation of $mf_{d-1}$.
Corollary~\ref{cor:bridge} gives an integer representation of
$mf_d$. Thus the positive integer multipliers are precisely the
positive elements of
$G_d\Z\cap E_{d-1}\Z$, proving \eqref{eq:multiplier-lcm}.
\end{proof}

\section{The exact coordinate-facet index}\label{sec:coordinate}

Throughout this section, $V=\{0,\ldots,6\}$ and $\Gamma=\Gamma_V$.
Let $T_0$ consist of all $21$ roots $r_{ij}$, and let
$T_0\subseteq T\subseteq\Gamma$ be any set of integral directions.
For $S\subseteq V$, write $C_q(S;R)$ for chains with base and
directions supported in $S$, and set
\begin{equation}\label{eq:coordinate-complex}
 A_q=\sum_{v\in V}C_q(V\setminus\{v\};R),\qquad b=d\Delta_V.
\end{equation}
The support condition is imposed on each generator separately.
Different summands of a chain in $A$ may collectively involve all seven
labels.

\begin{theorem}[Coordinate-facet image]\label{thm:index}
For every such $T$,
\begin{equation}\label{eq:index}
 \rho Z_4(A;\Z)=6\Z b,
 \qquad \rho Z_4(A;\D)=3\D b.
\end{equation}
Here $Z_4$ denotes the subgroup of closed chains. These identities
concern the coordinate-facet complex \eqref{eq:coordinate-complex};
they do not give a bound for unrestricted two-hidden-layer network
representations.
\end{theorem}

\subsection{The edge coefficient and its geometric normalization}

Consider the augmented \v{C}ech double complex
\begin{equation}\label{eq:cech}
 B_{p,q}=\bigoplus_{|I|=p+1}C_q(V\setminus I;R),
 \qquad D=\delta+(-1)^p d.
\end{equation}
The differential $\delta$ deletes elements of the increasingly ordered
omitted set $I$, with alternating signs. For each cube in the
coordinate-facet complex, the facets containing it form a nonempty
simplex. Exactness of the corresponding augmented simplicial
complexes, summed over all cubes, shows that the horizontal
augmentation resolves $A$.

Apply the smart vertical truncation
\[
 (\tau C)_q=C_q\ (q\geq2),\qquad
 (\tau C)_1=\ker(d:C_1\to C_0),\qquad (\tau C)_0=0.
\]
The quotient complex has vertical homology only in degree zero, equal to
$R$ for every $S$, including the empty set. Its horizontal complex
is the full six-simplex complex and has no positive homology.
Consequently $H_4(\Tot(\tau B))\cong H_4(A)$.
Every class has a representative with components
$z_{31},-z_{22},-z_{13},z_{04}$ satisfying
\begin{equation}\label{eq:descent}
 dz_{31}=0,\quad dz_{22}=\delta z_{31},\quad
 dz_{13}=\delta z_{22},\quad dz_{04}=\delta z_{13}.
\end{equation}
The horizontal augmentation of $z_{04}$ represents the class in $A$.

Lemma~\ref{lem:koszul} identifies the vertical homology with
$\Lambda^qW_T(S)$. In the degree-one row its horizontal third
homology is $R\omega$, with primitive generator
\begin{equation}\label{eq:omega}
 \omega=\sum_{|S|=3}(-1)^{\sum S}\tau_S,
 \qquad \tau_{abc}=e_{r_{ab}}+e_{r_{bc}}-e_{r_{ac}}.
\end{equation}
This normalization follows from the exact sequences of diagrams
\begin{align*}
 0&\longrightarrow W_T(S)\longrightarrow R^{(T|_S)}
       \longrightarrow\Gamma_S\otimes R\longrightarrow0,\\
 0&\longrightarrow\Gamma_S\otimes R\longrightarrow R^S
       \longrightarrow K(S)\longrightarrow0,
\end{align*}
where $K(S)=R$ for nonempty $S$ and $K(\varnothing)=0$.
The middle diagrams decompose into simplex complexes and hence have
no positive horizontal homology. The diagram $K$ has
$H_5(K)=R$; the two connecting maps give $H_3(W_T)=R$.
Applied to the primitive simplex boundary, these maps give the
triangle sum \eqref{eq:omega} without any additional scalar.
Let $c$ be the coefficient of $\omega$ represented by $z_{31}$.

\begin{lemma}[Normalization]\label{lem:normalization}
If $x$ is the augmentation of the last component of
\eqref{eq:descent}, then $\rho x=c b$. Classes with zero bottom
coefficient have zero coning image.
\end{lemma}

\begin{proof}
The standard closed triangle
\[
 L_{abc}=(0;r_{ab})+(r_{ab};r_{bc})-(0;r_{ac})
\]
has $\rho L_{abc}=-\Delta_{abc}$. Only its middle edge cones
nondegenerately, with orientation
$r_{bc}\wedge r_{ab}=-r_{ab}\wedge r_{ac}$.
Put $h_k(S)=(-1)^{\sum S}\Delta_S$ for $|S|=k$.

The difference between the bottom homology coordinates and
$c\omega$ is a horizontal boundary lifted from closed degree-one
chains in two-vertex spaces. Their coning images vanish by
dimension. The remaining discrepancy is a vertical boundary of
degree-two chains in three-vertex spaces, whose coning images also
vanish. The chain-map identity in degree two therefore gives
$\rho z_{31}=-c h_3$. No chain-map identity in degree one is used.

The chosen signs satisfy $\delta h_k=dh_{k+1}$. Indeed, if
$T'=S\cup\{v\}$ and $v$ has position $j$ in $T'$, its position
in $V\setminus S$ is $v-j$. Thus
$(-1)^{\sum S}(-1)^{v-j}=(-1)^{\sum T'}(-1)^j$.
Apply \eqref{eq:geometric-input} to \eqref{eq:descent} and use
top-boundary injectivity successively in dimensions three, four,
and five. This gives
\[
 \rho z_{22}=-c h_4,\qquad \rho z_{13}=-c h_5,
 \qquad \rho z_{04}=-c h_6.
\]
Since $\sum_{v\in V}v=21$, we have
$\sum_{|S|=6}h_6(S)=-d\Delta_V$, proving the formula.
Finally, $\rho(A_5)=0$ by dimension, so \eqref{eq:geometric-input}
shows that coning of closed $A_4$ chains descends to homology.
Thus the preceding changes of representative preserve the coning image.
\end{proof}

\subsection{Two finite obstruction calculations}

For root directions, choose the basis
$\{\tau_{s_0ij}:s_0<i<j,\ i,j\in S\}$ of $W_{T_0}(S)$,
where $s_0=\min S$. The inclusion maps have integer matrices, and
their exterior powers are computed by minors. In these bases, the
following proposition is certified by finite computations over the
integers. The complete sparse certificates and their verifiers are
specified in Section~\ref{app:certificates}.

\begin{proposition}[Certified differentials]\label{prop:finite-differentials}
In the vertical-first spectral sequence of \eqref{eq:cech} for root
directions, $d_2(\omega)$ has exact order two. A representative of
$d_3(2\omega)$ has exact order three in the horizontal
degree-zero homology of the $q=3$ row. There is an integral total
cycle with bottom coefficient six.
\end{proposition}

\begin{proof}[Finite-certificate proof]
The standard triangle lift has $105$ terms and represents
$\omega$. An integral root-star contraction fills its horizontal
boundary by a $105$-term $z_{22}$. Projecting
$\delta z_{22}$ to vertical homology gives $\beta_2$, a horizontal
cycle in
$\bigoplus_{|S|=5}\Lambda^2W(S)$, a free group of rank $315$.
The incoming inclusion matrix $M_2$ has $105$ columns. The certificate
specifies an integer vector $\lambda_2$ and a functional $\eta$ over $\F_2$
satisfying
\begin{equation}\label{eq:mod2-cert}
 \begin{aligned}
 M_2\lambda_2&=2\beta_2,\qquad \delta\beta_2=0
       &&\text{over }\Z,\\
 \eta M_2&=0,\qquad \eta\beta_2=1
       &&\text{over }\F_2.
 \end{aligned}
\end{equation}
All three nonzero entries of $\eta$ are on $S=\{0,1,2,3,4\}$:
\[
 \eta(\tau_{012}\wedge\tau_{034})
 =\eta(\tau_{013}\wedge\tau_{024})
 =\eta(\tau_{014}\wedge\tau_{023})=1.
\]
The cancellation of the incoming columns also has a geometric interpretation. Identify
$W(S)\otimes\F_2$ with $\Lambda^2\F_2^4$. Exterior multiplication
of two $2$-forms gives an alternating pairing with values in
$\Lambda^4\F_2^4$. Every four-vertex face lies in the exterior
square of a three-dimensional subspace; for the face omitting zero
this is the kernel of the coordinate-sum functional. The fourth
exterior product vanishes in every case. The supplied integer vectors
verify the remaining equalities in \eqref{eq:mod2-cert}, proving
exact order two in the horizontal homology group.

For $2\omega$, the next certified descent has obstruction
$\beta_3\in\bigoplus_{|S|=6}\Lambda^3W(S)$, with $840$
coordinates. The incoming matrix $M_3$ has $420$ columns. An
integer vector $\lambda_3$ and a functional $\psi$ over $\F_3$
satisfy
\begin{equation}\label{eq:mod3-cert}
 \begin{aligned}
 M_3\lambda_3&=3\beta_3 &&\text{over }\Z,\\
 \psi M_3&=0,\qquad \psi\beta_3=1 &&\text{over }\F_3.
 \end{aligned}
\end{equation}
The functional is supported on $S=\{0,1,2,3,4,5\}$.
Writing $ij\mid kl\mid mn$ for
$\tau_{0ij}\wedge\tau_{0kl}\wedge\tau_{0mn}$, its six values are
\begin{center}
\begin{tabular}{@{}crcr@{}}
\toprule
Exterior triple & $\psi$ & Exterior triple & $\psi$\\
\midrule
$12\mid13\mid45$ & $-1$ & $13\mid14\mid25$ & $-1$\\
$12\mid14\mid35$ & $ 1$ & $13\mid15\mid24$ & $ 1$\\
$12\mid15\mid34$ & $-1$ & $14\mid15\mid23$ & $-1$\\
\bottomrule
\end{tabular}
\end{center}
All other values are zero. Equations \eqref{eq:mod3-cert} prove
exact order three in the stated horizontal cokernel.

For bottom coefficient six, the successive term counts are
$105,805,5462,21104$. Substitution into
\eqref{eq:cubical-boundary}, together with the signed inclusion maps,
verifies every equation in \eqref{eq:descent}. Its horizontal augmentation
is an integral closed chain $X$ with $17{,}217$ nonzero terms.
Lemma~\ref{lem:normalization} gives
\begin{equation}\label{eq:coordinate-lift}
 dX=0,\qquad \rho X=6b.
\end{equation}
The exterior matrices and boundaries are reconstructed from the
specified bases and checked exactly over the integers.
\end{proof}

\subsection{Dilation and the prime support of higher differentials}

The doubling operation of \cite[Sections 4 and 5.3]{brunck} extends
integrally to every positive integer scale. Considering all scales
gives a sharper torsion constraint in the coordinate-facet \v{C}ech
spectral sequence. These annihilator bounds do not give lower bounds
for unrestricted networks.

\begin{proposition}[Dilation annihilators]\label{prop:dilation-annihilators}
Let $E^r_{p,q}$ be the vertical-first spectral sequence of the smartly
truncated double complex \eqref{eq:cech}, over $\Z$ or a localization of
$\Z$, for any root-containing set of integral directions. For $q\geq1$
and $r\geq2$, put
\[
 G_{q,r}=\gcd_{m\geq2}\bigl(m^{q+r-1}-m^q\bigr).
\]
Then
\begin{equation}\label{eq:dilation-annihilator}
 G_{q,r}\,d_r=0,
 \qquad d_r:E^r_{p,q}\longrightarrow E^r_{p-r,q+r-1}.
\end{equation}
Writing $s=r-1$, the integer in \eqref{eq:dilation-annihilator} is
\begin{equation}\label{eq:fixed-divisor-formula}
 G_{q,r}
 =2^{\min\{q,a_2(s)\}}
   \prod_{\substack{p\text{ odd prime}\\p-1\mid s}}
       p^{\min\{q,1+v_p(s)\}},
 \qquad
 a_2(s)=
 \begin{cases}
  1,&s\text{ odd},\\
  2+v_2(s),&s\text{ even}.
 \end{cases}
\end{equation}
In particular, every $d_2$ is annihilated by $2$, and every even-page
higher differential vanishes over $\D$. A nonzero $p$-primary part of a
higher differential is possible only if $p-1$ divides $r-1$.
\end{proposition}

\begin{proof}
For $m\geq2$, define dilation followed by subdivision on a cubical
generator by
\begin{equation}\label{eq:all-scale-subdivision}
 S_m(\gamma;t_1,\ldots,t_q)
 =\sum_{0\leq j_1,\ldots,j_q<m}
   (m\gamma+j_1t_1+\cdots+j_qt_q;t_1,\ldots,t_q).
\end{equation}
For $q=0$, this means $S_m(\gamma)=m\gamma$ as a base point, with
coefficient one. In each direction, the boundary of the sum telescopes
to the two subdivided outer faces. Hence $dS_m=S_md$. Formula
\eqref{eq:all-scale-subdivision} preserves coordinate supports and
commutes with their inclusions, so it induces a map of the double
complex and of its smart truncation.

Augmentation gives $\epsilon S_m=m^q\epsilon$ in degree $q$.
By Lemma~\ref{lem:koszul}, $S_m$ acts as multiplication by $m^q$ on
each summand of $E^1_{p,q}$. It acts by the same scalar on every later
$E^r_{p,q}$, since each successive page is obtained from kernels and
quotients in the same bidegree. Naturality of $d_r$ gives
\[
 m^{q+r-1}d_r=d_r m^q=m^q d_r.
\]
The ideal of $\Z$ generated by these scalar differences is generated by
$G_{q,r}$; a finite B\'ezout combination therefore proves
\eqref{eq:dilation-annihilator}. No identification of a later-page
target with its $E^2$ presentation is required.

To compute this greatest common divisor, consider its valuation at each prime.
For a prime $p$ and $m$ divisible by $p$,
\[
 v_p\bigl(m^q(m^s-1)\bigr)=qv_p(m),
\]
whose minimum is $q$. For odd $p$, if $p-1$ does not divide $s$,
a generator of $(\Z/p\Z)^\times$ gives a value not divisible by $p$.
If $p-1\mid s$, the least valuation for $m$ prime to $p$ is
$1+v_p(s)$. Indeed, Fermat's theorem and the binomial valuation identity
$v_p(a^u-1)=v_p(a-1)+v_p(u)$ for $a\equiv1\pmod p$ give the lower
bound. To attain it, take a lift $m$ of a generator modulo $p$ with
$v_p(m^{p-1}-1)=1$. Such a lift exists: among the $p$ lifts modulo
$p^2$, precisely one satisfies $m^{p-1}\equiv1\pmod{p^2}$, because
the derivative $(p-1)m^{p-2}$ is nonzero modulo $p$.
Taking the minimum with $q$ gives the odd-prime factors.

For $p=2$ and odd $s$, an odd $m$ gives minimum valuation one,
attained at $m=3$. For even $s$ and odd $m$, factorization and the
same binomial identity give
\[
 v_2(m^s-1)=v_2(m-1)+v_2(m+1)+v_2(s)-1
            \geq2+v_2(s),
\]
with equality at $m=3$. Again take the minimum with $q$. This proves
\eqref{eq:fixed-divisor-formula}. When $r$ is even, $s$ is odd and
no odd prime has $p-1\mid s$, so $G_{q,r}=2$; the remaining
conclusions follow.
\end{proof}

\begin{remark}[Bottom-row bounds and their limits]
\label{rem:bottom-row-primes}
For $q=1$ the bound is squarefree:
\[
 G_{1,r}=\prod_{p-1\mid r-1}p.
\]
For $r=2,3,4,5,6,7$ these integers are $2,6,2,30,2,42$.
Thus, after inverting two, the first possible page for an obstruction
at an odd prime $p$ is $r=p$. The annihilator bounds do not assert
that any such obstruction is nonzero. Nor do they determine extension
orders in the abutment or remove the coordinate-support restriction.
\end{remark}

\subsection{Survival and arbitrary integral directions}

\begin{proof}[Proof of Theorem~\ref{thm:index}]
For root directions, the $E^2_{3,1}$ source is $\Z\omega$. There are no
incoming higher differentials to this source after smart truncation.
Proposition~\ref{prop:finite-differentials} makes its $E^3$
survival subgroup $2\Z\omega$.

The order-three class in \eqref{eq:mod3-cert} is unaffected by
the $d_2$ indeterminacy. Proposition~\ref{prop:dilation-annihilators}
shows that every incoming $d_2$ image is annihilated by two, so it
intersects the cyclic subgroup generated by the order-three class
trivially. Hence $d_3(2\omega)$ still has exact order three on $E^3$,
and the surviving source is $6\Z\omega$. No $d_r$ for $r>3$ can
leave column three. Over $\D$, every $d_2$ vanishes and the source
survival subgroup is $3\D\omega$.

For an enlarged direction set $T$, replace each additional degree-one direction
by an integral root path with the same endpoints, chosen within its
support. Extend multiplicatively in the cubical Koszul algebra and
fix the root generators. This is a support-preserving chain
retraction to the root complex and induces a retraction of the
\v{C}ech double complexes. It preserves the primitive bottom
coefficient. Thus every completed descent for $T$ produces a root
descent with the same coefficient, so that coefficient belongs to
$6\Z$, or to $3\D$ after localization. Commutation of the path
retraction with $\rho$ is not needed:
Lemma~\ref{lem:normalization} computes the geometric coefficient
separately in both complexes.

Lemma~\ref{lem:normalization} proves containment in the right
sides of \eqref{eq:index}. The cycle $X$ in
\eqref{eq:coordinate-lift} attains $6b$, and $X/2$ attains $3b$
over $\D$. This proves equality.
\end{proof}

\section{Integral representations of six- and seven-input maxima}
\label{sec:positive}
\subsection{An augmentation repair for the sixfold construction}
\label{sec:six-max-seven}

The coordinate lift $X$ has coning coefficient six, but its
augmentation need not vanish. Adjoining three integral directions
allows us to cancel the augmentation while preserving the coning
image. Work in
$\Gamma_V=\{a\in\Z^7:\sum_i a_i=0\}$, and set
\begin{align}\label{eq:three-repair-directions}
 t_1&=(-1,-1,0,1,1,0,0),\notag\\
 t_2&=(-1,-1,0,1,0,1,0),\\
 t_3&=(-1,0,-1,1,1,0,0).\notag
\end{align}
Let $T_+=\{r_{ij}:0\leq i<j\leq6\}\cup\{t_1,t_2,t_3\}$
and $W_+=\ker(\Z^{T_+}\to\Gamma_V)$. The six star roots form
a lattice basis, and $W_+$ has rank $18$.

\begin{proposition}[A coefficient-six augmentation repair]
\label{prop:six-augmentation-repair}
There is an integral degree-four chain
$Y_6\in C_4(\Gamma_V,T_+;\Z)$ satisfying
\[
 dY_6=0,\qquad\epsilon Y_6=0,\qquad\rho Y_6=6d\Delta_6.
\]
One explicit choice has $66{,}402$ nonzero cubical terms.
\end{proposition}

\begin{proof}[Finite-certificate proof]
Let $X$ be the $17{,}217$-term coordinate cycle with
$\rho X=6d\Delta_6$, and write $w=[X]\in\Lambda^4W_+$.
Take the root-star basis of
$W_{T_0}=\ker(\Z^{T_0}\to\Gamma_V)$, followed by the three relations
\[
 r_{t_j}=e_{t_j}-\sum_{i=1}^6(t_j)_i e_{r_{0i}},
 \qquad j=1,2,3,
\]
as an integral basis of $W_+$.

The certificate gives $803$ decomposable integral classes
\[
 z_\nu=r_{\nu,1}\wedge r_{\nu,2}\wedge
              r_{\nu,3}\wedge r_{\nu,4}\in\Lambda^4W_+.
\]
Each $r_{\nu,i}$ is an explicitly specified integral relation among
the $24$ directions. For each $\nu$, all directions occurring in
its four relations lie in a specified proper linear hyperplane
\[
 H_{A_\nu}=\left\{x\in\Gamma_V\otimes\R:
                       \sum_{i\in A_\nu}x_i=0\right\}.
\]
There are $42$ distinct hyperplanes in the complete certificate.
These classes satisfy the integer identity
\begin{equation}\label{eq:six-repair-exterior-identity}
 w-\sum_{\nu=1}^{803}z_\nu=2v,
 \qquad v\in\Lambda^4W_+.
\end{equation}
The vector $v$ has $1253$ nonzero coordinates and maximum
absolute coefficient $11$. The independent verifier checks that
every relation evaluates to zero in $\Gamma_V$, that every
direction belongs to its specified hyperplane, and that
\eqref{eq:six-repair-exterior-identity} holds in every integer
coordinate. It also verifies the corresponding equality in
$\Lambda^4\Z^{T_+}$ by exact coordinate arithmetic.

For an integral relation $r=\sum_t a_t e_t$ supported in a linear
space $H$, start at the origin and realize each displacement $a_t t$
by repeated oriented steps in the corresponding allowed direction.
Since $\sum_t a_t t=0$, this gives a closed integral
one-chain $L_r$ in $H$ with $\epsilon L_r=r$.
Products of these loops in the cubical Koszul algebra give closed
four-chains
\[
 \kappa_\nu=L_{r_{\nu,1}}L_{r_{\nu,2}}
             L_{r_{\nu,3}}L_{r_{\nu,4}}
\]
with homology coordinates $z_\nu$ and all base points and directions
in $H_{A_\nu}$. We have $\rho\kappa_\nu=0$ for each summand: in a
five-dimensional hyperplane this follows from
$d\rho\kappa_\nu=0$ and top-boundary injectivity, and in smaller
dimension every coning term is degenerate. Put
$K_H=\sum_\nu\kappa_\nu$.

Choose a closed integral four-chain $c_v$ realizing $v$ by
the same loop-product construction. By reflection,
\[
 \rho(c_v+J_4c_v)=0,
 \qquad\epsilon(c_v+J_4c_v)=2\epsilon c_v.
\]
Consequently
\begin{equation}\label{eq:six-repaired-cycle}
 Y_6=X-K_H-c_v-J_4c_v
\end{equation}
has zero boundary and zero augmentation, with
$\rho Y_6=6d\Delta_6$.

The explicit chains have $22{,}067$ terms in $K_H$ and
$24{,}319$ terms in each of $c_v$ and $J_4c_v$. Their combination
\eqref{eq:six-repaired-cycle} has $66{,}402$ terms after collecting
identical cubes. The standard-library checker verifies all
boundaries, the zero augmentation, and the support of every
hyperplane group exactly over the integers.
\end{proof}

\begin{theorem}[An integer sixfold seven-input maximum]
\label{thm:six-max-seven}
The function $6M_7$ has a finite-width integer network with two
hidden ReLU layers. In particular,
\begin{equation}\label{eq:improved-E6}
 E_6\in\{2,6\}.
\end{equation}
\end{theorem}

\begin{proof}
By Lemma~\ref{lem:koszul}, the augmentation-zero cycle of
Proposition~\ref{prop:six-augmentation-repair} has an integral
filling $U_6\in C_5(\Gamma_V,T_+;\Z)$ with $dU_6=Y_6$.
The supplementary data specify such a filling, with
$376{,}126$ nonzero terms. A separate standard-library verifier
reconstructs its $3{,}761{,}260$ signed boundary faces and checks
the identity $dU_6=Y_6$; its coefficients have absolute value
at most $144$ and its base coordinates have absolute value at most
$11$. Therefore
\[
 d\rho U_6=\rho dU_6=6d\Delta_6,
 \qquad\rho U_6=6\Delta_6,
\]
where the second equality uses top-boundary injectivity in the
six-dimensional ambient space. Passing from $\rho U_6$ to ordinary
support functions gives an integer sum of paramid supports modulo
$\La_5$. Coefficient-preserving proper-face completion removes
this remainder: each proper simplex face has at most six vertices,
and the known integer $6M_6$ identity represents six times its
support function after an integral substitution. The proper faces
of the paramids already have integer two-hidden-layer representations.
Thus Corollary~\ref{cor:bridge} supplies an integer representation
of $6f_6$, and the translation identity supplies $6M_7$.
This argument is independent of the construction of $2M_6$ below.

The missing-prime lower bound \cite[Theorem 2]{ahm} excludes odd
multipliers of $M_7$. Since the admissible integer multipliers form an ideal,
$2\mid E_6\mid6$, proving \eqref{eq:improved-E6}.
\end{proof}

Thus the two-adic valuation of the least multiplier is
$v_2(E_6)=1$. The remaining question is whether the factor three
is necessary. Equivalently, dyadic two-hidden-layer representability
of $M_7$ amounts to deciding between $E_6=2$ and $E_6=6$.
The coefficient-six construction is certified by a finite integral
augmentation repair and the constructive filling and completion
lemmas; the full network weight arrays have not been expanded.

\subsection{The six-input exponent}

The five-dimensional certificate uses an integral source chain in
the rational polytope complex and requires no integral saturation
assertion.

\begin{proposition}[Five-dimensional simplex certificate]
\label{prop:five-simplex}
For the six-vertex simplex there is an integral degree-four chain
$u$ with $43{,}623$ terms and $\rho u=4\Delta_5$.
\end{proposition}

\begin{proof}[Finite-certificate proof]
The starting identity is Schroeder's integer $2M_5$ formula
\cite[Equation (4.1), Table 2]{schroeder}. A six-cube conversion
of its three full-dimensional terms is given in
Section~\ref{app:six-cube}; it gives an integral lift of
$2\Delta_4$. The specified combination of these facet lifts and
degenerate corrective cubes is a closed $874$-term chain $x$ with
\[
 dx=0,\qquad\rho x=2d\Delta_5.
\]
The checker reconstructs each facet lift, verifies the total signed
mass on each facet, and checks the degeneracy of every corrective
cube through its minors.

The augmentation repair supplies a closed chain $\kappa_H$, a
sum of closed chains in four-dimensional rational direction
spaces, and a closed chain $c_w$. Closedness gives
$d\rho\kappa_H=0$; top-boundary injectivity, applied separately
in the support of each summand, gives
$\rho\kappa_H=0$. Let $S_2$ be twofold cubical dilation followed
by subdivision. The certificate verifies
\begin{equation}\label{eq:five-repair}
 y=2x-2\kappa_H-S_2c_w+2c_w,
 \qquad dy=0,\qquad\epsilon y=0.
\end{equation}
Coning followed by subdivision is compatible with positive dilation,
so $\rho S_2c_w=2\rho c_w$. Hence $\rho y=4d\Delta_5$.
No vanishing assumption on $\rho c_w$ is needed. The explicit
$43{,}623$-term chain $u$ satisfies $du=y$, as verified from its
$348{,}984$ signed boundary faces. Equation
\eqref{eq:geometric-input} gives
$d\rho u=4d\Delta_5$ and then $\rho u=4\Delta_5$.
All these identities hold in the rational polytope complex, with
integral source chains throughout.
\end{proof}

\begin{proof}[Proof of Theorem~\ref{thm:main}]
The $6M_7$ construction is Theorem~\ref{thm:six-max-seven}. Apply
Corollary~\ref{cor:bridge} to Proposition~\ref{prop:five-simplex}.
The proper simplex faces have at most five vertices, so the known
integer $2M_5$ identity, scaled by two, represents their supports
with the retained multiplier four. Thus $4M_6$ has an integer
representation. Subtracting this network from the known integer $6M_6$ network gives an integer
representation of $2M_6$.

For either $n=6$ or $n=7$, suppose an odd integer $m$ were a
multiplier for a two-hidden-layer integer network. Dividing the
output by $m$ would represent
$M_n$ over $\Z[1/m]$. The missing-prime bound of
\cite[Theorem 2]{ahm}, with prime two, requires at least
$\lceil\log_2 n\rceil=3$ hidden layers. This contradiction proves
that both least multipliers are even. The constructed multipliers
give $E_5\mid2$ and $E_6\mid6$, as required.
\end{proof}

The exact coordinate index six does not imply $6\mid E_6$.
The exponent is defined using all networks, whereas
Theorem~\ref{thm:index} concerns closed chains in $A$.
Whether $3\mid E_6$ therefore remains open.

\section{A relative kernel formulation of the remaining problem}
\label{sec:relative}

The coordinate-facet index yields a relative kernel class whose image
in the full direction complex detects dyadic representability.
Work over $\D$, and let $K_q=\ker(\rho:C_q\to P_{q+1})$ for
$q\geq2$, with zero differential out of degree two. By the chain-map
identity \eqref{eq:geometric-input}, this is a complex in the
degrees relevant to $H_3$ and $H_4$. Put $K_A=K\cap A$.
Enlarge $T$ to include the directions in the seven facet embeddings
of the chain $u$ of Proposition~\ref{prop:five-simplex}, and set
\begin{equation}\label{eq:relative-a}
 a=\frac14\sum_{v=0}^6(-1)^v\iota_v(u),\qquad \rho a=b.
\end{equation}
Here $\iota_v$ uses the increasing order on the facet opposite $v$.
Although $a$ need not be closed, $\rho(da)=0$, which defines the class
$\theta=[da]\in H_3(K_A;\D)$.

\begin{proposition}[A relative order-three class]\label{prop:theta}
The annihilator of $\theta$ is $3\D$, and the following statements
are equivalent:
\begin{enumerate}
\item $\theta$ maps to zero in $H_3(K;\D)$;
\item there is a closed $x\in C_4(\Gamma,T;\D)$ with
$\rho x=b$;
\item $\theta$ lies in the image of the connecting homomorphism
$H_4(K/K_A;\D)\to H_3(K_A;\D)$.
\end{enumerate}
These equivalent conditions imply a dyadic two-hidden-layer
representation of $M_7$.
\end{proposition}

\begin{proof}
The chain $3a-X/2$ belongs to $K_A$ and has boundary $3da$, so
$3\theta=0$. If $r\theta=0$, take $k\in K_{A,4}$ with
$dk=rda$. Then $ra-k$ is a closed coordinate lift of $rb$;
Theorem~\ref{thm:index} forces $r\in3\D$. This proves the
annihilator assertion. The equivalence of the first two statements
follows from $x=a-k$ and $k=a-x$. The equivalence with the third
follows from the long exact homology sequence of the pair of complexes.

If such an $x$ exists, replace it by $(x-J_4x)/2$.
Lemma~\ref{lem:reflection} preserves its coning image and cancels
its augmentation. Lemma~\ref{lem:koszul} supplies a dyadic filling,
and top-boundary injectivity gives a dyadic simplex certificate.
Clear the denominators of the filling coefficients by a power of two,
chosen to be at least two. The retained multiplier on every proper
simplex face is then divisible by $E_5=2$. Corollary~\ref{cor:bridge}
gives an integer representation of that power of two times $f_6$;
division at the final output gives the required dyadic network.
\end{proof}

Changing the coordinate preimage $a$ changes $da$ by a boundary
in $K_A$, so the class is independent of the choice. The certificates
also give the integral target $t=4da$ with $48{,}234$ terms in
$34$ directions, and verify
\[
 d(4a)=t,\quad dt=0,\qquad
 k_3=3(4a)-2X\in K_A,\quad dk_3=3t.
\]
The class $[t]$ has order three over $\Z$: if it bounded
in $K_A$, it would repair $4a$ to a coordinate lift of $4b$,
contradicting the index six.

\begin{theorem}[A complete global dyadic obstruction]
\label{thm:global-theta}
In dimension six, take all nonzero lattice directions and write
$K^{\mathrm{all}}=\ker\rho$ for the corresponding complex in
degrees at least two. Let $\theta_\infty$ be the image of the
coordinate class $\theta$ in
$H_3(K^{\mathrm{all}};\D)$. The following are equivalent:
\begin{enumerate}
\item $\theta_\infty=0$;
\item $M_7$ has a dyadic two-hidden-layer representation;
\item $E_6=2$.
\end{enumerate}
If these statements fail, then $E_6=6$ and
$\operatorname{Ann}_{\D}(\theta_\infty)=3\D$.
Moreover, $\theta_\infty=0$ if and only if the image of $\theta$ in
$H_3(K(\Gamma,T');\D)$ vanishes for some finite root-containing
direction set $T'$ containing the directions of $a$.
\end{theorem}

\begin{proof}
The proof of Proposition~\ref{prop:theta} applies to finite chains
in the full direction complex and shows that the first statement
implies the second. For the converse, restrict a dyadic network
for $M_7$ to obtain one for $f_6$. Positive denominator clearing
gives an integer network for $2^s f_6$ for some $s\ge0$.
Lemma~\ref{lem:integral-paramid-normal-form} gives an integral
$u\in C_5^{\mathrm{all}}(\Z^6;\Z)$ with
$\rho u=2^s\Delta_6$. The chain
$x=2^{-s}du$ is closed and satisfies $\rho x=b$.
Thus $a-x\in K_4^{\mathrm{all}}$ and $d(a-x)=da$, proving
$\theta_\infty=0$.

A dyadic network exists exactly when some power of two is an
integer multiplier, or equivalently when $E_6$ is a power of two.
Theorem~\ref{thm:main} restricts $E_6$ to $\{2,6\}$,
which proves the stated alternatives. When these equivalent conditions
hold, divide the output of an integer $2M_7$ network by two. This gives
integral hidden-layer arrays and half-integral output coefficients.
The image $\theta_\infty$
is annihilated by three. Since $\D/3\D\cong\mathbb F_3$,
its annihilator is $3\D$ whenever it is nonzero.
Finally, every chain witnessing a boundary uses only finitely
many lattice directions. Such a witness lies in $K(\Gamma,T')$
for a finite enlargement $T'$ as stated, and the converse follows
from inclusion in $K^{\mathrm{all}}$.
\end{proof}

The full direction set is essential in this necessity statement.
For a fixed finite direction set, Proposition~\ref{prop:theta}
continues to characterize closed cubical lifts in that complex;
an arbitrary network may require additional directions.
The resolved augmentation class is treated separately in
Section~\ref{next:global-two-primary-resolved} of the supplement.

\subsection{An integral target for the global dyadic problem}

Let $T$ contain the directions of $a$ and $X$, and write
$K_q(T;\Z)=\ker(\rho:C_q(\Gamma,T;\Z)\to P_{q+1})$ in degrees
at least two. For any localization $R$ of $\Z$, finite-support
denominator clearing gives
$K(T;R)=K(T;\Z)\otimes_{\Z}R$. The finite chains used below are
\begin{equation}\label{next:three-target}
 t=4da,\qquad k_3=12a-2X\in K_4(T;\Z),
 \qquad dk_3=3t.
\end{equation}
In particular, $t\in K_3(T;\Z)$ is closed and its homology class
is annihilated by three. Set
$\Z_{(3)}=\{u/v\in\Q:3\nmid v\}$.

\begin{proposition}[Integral descent of the order-three target]
\label{next:integral-target}
For every such direction set $T$, the following conditions are equivalent:
\begin{enumerate}
\item $t=dk$ for a finite $k\in K_4(T;\Z)$;
\item $t=dk$ for a finite $k\in K_4(T;\Z[1/2])$;
\item $t=dk$ for a finite $k\in K_4(T;\Z_{(3)})$.
\end{enumerate}
For the full set of nonzero lattice directions, these conditions are
also equivalent to dyadic two-hidden-layer representability of $M_7$.
Thus the global dyadic problem amounts to deciding whether the fixed
integral target $t$ has an integral kernel filling.
\end{proposition}

\begin{proof}
It remains to show that the third condition implies the first. Choose a
positive common denominator $n$ of the finitely many coefficients of
$k$, with $3\nmid n$. Then $\ell=nk\in K_4(T;\Z)$ and
$d\ell=nt$. Choose integers $u,v$ with $un+3v=1$. The chain
\begin{equation}\label{next:bezout-integral-filler}
 k_{\Z}=u\ell+vk_3
\end{equation}
is integral, has zero coning image, and satisfies $dk_{\Z}=t$.
The other implications follow from inclusion of coefficient rings.

The image of $\theta=[da]$ in the global dyadic kernel homology
vanishes if and only if the class of $t=4da$ vanishes, because four
is a unit. Apply Theorem~\ref{thm:global-theta} and the
preceding argument.
\end{proof}

\begin{remark}[Constructing a bounded multiplier from a witness]
\label{next:construct-from-integral}
If $k_{\Z}$ is an integral witness, then
\[
 x_2=X-4a+k_{\Z}
\]
is a closed integral chain with $\rho x_2=2b$. Reflection gives the
integral, augmentation-zero cycle $x_2-J_4x_2$ with coning image $4b$.
An integral Koszul filling and coefficient-preserving face completion
therefore construct an integer network for $4M_7$, since every proper
simplex face retains the multiplier four and $E_5=2$.
Subtract this network from the integer $6M_7$ network of
Theorem~\ref{thm:six-max-seven} to obtain $2M_7$. Thus an integral
filling witness would determine the exact exponent $E_6=2$.
\end{remark}

The restriction on denominators is essential. The rational chain
$k_3/3$ always fills $t$, so rational solvability does not decide
the integral filling problem. Proposition~\ref{next:integral-target} applies unchanged
to any finite chain dictionary containing the support of $k_3$.

\section{Finite arithmetic separation and top-class searches}
\label{sec:finite-separation}

\subsection{Arithmetic separation in a finite pair-max dictionary}
\label{subsec:finite-separation}

Label the coordinates by $0,\ldots,6$, and set
\[
 \mathcal E=\{(i,j):0\leq i\leq j\leq6\},\qquad
 m_{ij}(x)=\max(x_i,x_j),\qquad m(x)=(m_{ij}(x))_{(i,j)\in\mathcal E}.
\]
Thus the $28$ features include $m_{ii}(x)=x_i$.
For multisets $A,B$ of three pairs, let $a_{A,B}\in\Z^{\mathcal E}$
be the difference of their pair-count vectors. If $a_{A,B}\ne0$,
divide it by the positive gcd of its entries and choose the sign with
positive first nonzero entry in lexicographic order on $\mathcal E$;
call the resulting row $g_{A,B}$.
Consider three families of pairs: the $57$ pairs in the public
seven-input certificate of \cite{ruess}, the $109$ pairs listed in
\cite[Appendix A]{wangbasu}, and all pairs of edge-disjoint, loop-free,
simple three-edge forests on the common seven-vertex set. Isolated
vertices are allowed; edge disjointness means $A\cap B=\varnothing$.
Apply every coordinate permutation in $S_7$ to these pairs, normalize the
resulting rows as above, and let $G_{\mathrm{nat}}$ be the set of distinct rows.
For a unital subring $R\subseteq\Q$, put
\begin{equation}\label{eq:natural-dictionary}
 \mathcal D_{\mathrm{nat},R}
 =\operatorname{span}_R\bigl(
   \{m_{ij}:(i,j)\in\mathcal E\}
   \cup\{\sigma(g\cdot m):g\in G_{\mathrm{nat}}\}\bigr).
\end{equation}
This includes the original maximum-of-two-sums gates because
\[
 \max\Bigl\{\sum_{e\in A}m_e,\sum_{e\in B}m_e\Bigr\}
 =\sum_{e\in B}m_e+\sigma(a_{A,B}\cdot m),\qquad
 \sigma(-g\cdot m)=\sigma(g\cdot m)-g\cdot m.
\]
Primitive normalization may enlarge the integer module, so an obstruction
for \eqref{eq:natural-dictionary} also applies to the unnormalized gates.
The coefficients are unrestricted by symmetry.

\begin{theorem}[Finite-dictionary arithmetic separation]
\label{thm:finite-separation}
For $R=\Z_{(3)}=\{a/b\in\Q:3\nmid b\}$,
\[
 M_7\in\mathcal D_{\mathrm{nat},\Q},\qquad
 M_7\notin\mathcal D_{\mathrm{nat},R}.
\]
Moreover, for $m\in\Z$, $mM_7\in\mathcal D_{\mathrm{nat},\Z}$
implies $3\mid m$.
These assertions allow arbitrary nonsymmetric coefficients and all
$28$ auxiliary functions in \eqref{eq:natural-dictionary}.
\end{theorem}

\begin{proof}
The public $57$-seed identity proves the rational assertion. The
supplement verifies it on a full braid cone by exact hinge cancellation;
the permutation sum gives the identity on the whole input space.
The gate identities above place it in $\mathcal D_{\mathrm{nat},\Q}$.

For the exclusion, let
$H=\operatorname{AGL}(1,5)\times S_2\leq S_7$: the affine group
acts on coordinates $0,\ldots,4$, identified with $\F_5$, and the
second factor interchanges $5,6$. Thus $|H|=40$.
The supplementary certificate supplies $4194$ integer points $x_j$
and coefficients $\lambda_j\in\F_3$. On functions taking values
in $R$ at integer points, define
\begin{equation}\label{eq:natural-character}
 \ell(f)=\sum_j\lambda_j\sum_{h\in H}f(hx_j)\pmod3,
\end{equation}
where reduction uses $R/3R\cong\F_3$.
The certificate verifies
\[
 \ell(m_{ij})=0,\qquad
 \ell\bigl(\sigma(g\cdot m)\bigr)=0\quad(g\in G_{\mathrm{nat}}),
 \qquad \ell(M_7)=1.
\]
The verification programs reconstruct both public seed lists and exhaustively
enumerate the forest pairs. They establish containment in an explicitly
listed certificate set of $804{,}202$ primitive rows, a superset of
$G_{\mathrm{nat}}$, and check its complete coverage by $20{,}992$
$H$-orbits modulo sign. Each representative is evaluated using the
full $40$-term sum in \eqref{eq:natural-character}; all $28$ auxiliary
pairings are checked separately. Invariance of $\ell$ and the signed
gate identity above extend these checks to every row and both signs.
The resulting identities show that the functional annihilates each
generator, hence the whole $R$-module. Its value on $M_7$ excludes an
$R$-representation, and its value on $mM_7$ gives the divisibility
assertion.
\end{proof}

The theorem concerns the specified dictionary, which need not contain
all degree-three pair-max gates. To compare this obstruction with
unrestricted representations, we must account for the omitted
lower-dimensional corrections.

\subsection{Lower-dimensional remainders and the top quotient}
\label{subsec:top-criterion}

Work on the reduced input space $\R^6$ for $f_6$. Let $\mathcal V_\Z$
be the integer span of lattice-polytope support functions, put
$\mathcal V_R=R\mathcal V_\Z$ for $R=\Z_{(3)}$, and let
$\mathcal N_R$ be the additive group of homogeneous functions with
two-hidden-layer representations over $R$. Define $\mathcal N_\Z$
analogously, replacing $R$ by $\Z$. Write
\[
 \pi:\mathcal V_R\longrightarrow
       \mathcal V_R/(\mathcal V_R\cap\La_5).
\]
Here $\La_5$ is the real lower-dimensional support space. An integral
decomposition of an element in this intersection requires the completion
lemma.

\begin{lemma}[Integral lower-dimensional remainders]
\label{lem:proper-remainder}
If $F\in\mathcal V_\Z\cap\La_5$, then $2F\in\mathcal N_\Z$.
Consequently $\mathcal V_R\cap\La_5\subseteq\mathcal N_R$.
\end{lemma}

\begin{proof}
Choose a finite integer support-function presentation of $F$ and
separate the terms of dimension at most five. The remaining sum
still belongs to $\La_5$. Lemma~\ref{lem:completion} expresses it
as an integer linear form and an integer combination of proper
lattice faces. Thus all remaining polytopes have dimension at most
five, with integer coefficients.

Triangulate each using its lattice vertices. The support-function
valuation identity, including nonempty intersection faces, gives an integer
combination of simplex supports with at most six vertices.
Twice each support is an integral input substitution into the
established representation of $2M_6$, repeating vertices when
necessary. The same factor two handles every term; signed ReLU
pairs carry the integer linear remainder. This proves the first
assertion. For the second, clear a common coefficient denominator
$n$ prime to three. Apply the first assertion to $nF$ and divide
the final output coefficients by the unit $2n\in R$.
\end{proof}

\begin{proposition}[A sufficient top-class criterion]
\label{prop:top-criterion}
Let $g_1,\ldots,g_N\in\mathcal V_\Z\cap\mathcal N_\Z$ be any
fixed finite collection. If
\[
 \pi(f_6)=\sum_{j=1}^N r_j\pi(g_j),\qquad r_j\in\Z_{(3)},
\]
then $E_6=2$.
\end{proposition}

\begin{proof}
Choose $n>0$, prime to three, with all $nr_j\in\Z$.
The remainder $nf_6-\sum_jnr_jg_j$ belongs to
$\mathcal V_\Z\cap\La_5$. The lemma therefore constructs an
integer representation of $2nf_6$, hence of $2nM_7$.
Combine it with the established integer $6M_7$ representation
using Bezout coefficients. Since $\gcd(2n,6)=2$, this gives
$2M_7$; the parity lower bound proves $E_6=2$.
\end{proof}

To obstruct the quotient criterion, a separating character must also
annihilate $\mathcal V_R\cap\La_5$.
The character in Theorem~\ref{thm:finite-separation} does not: the
supplement verifies its nonzero value on
$q_* =\sigma(m_{45}+2x_6-m_{13}-m_{16}-m_{23})$ and proves,
after setting $x_0=0$, that $q_*\in\mathcal V_\Z\cap\La_5$.
It therefore does not descend to the top-dimensional quotient and cannot
distinguish $E_6=2$ from $E_6=6$.
\section{Discussion and reproducibility}\label{sec:discussion}

The integral normal form and coefficient-preserving completion reduce
the multiplier problem to arithmetic in polytope and chain modules.
For six inputs, they determine the least multiplier. For seven inputs,
the remaining question concerns the prime three. The coordinate-facet
index and the finite character detect this prime in two restricted
modules; neither establishes a global obstruction. The next step is
therefore to decide the vanishing of the class in
Theorem~\ref{thm:global-theta}. A finite integral kernel filling would
prove $E_6=2$, while nonvanishing of the global class would prove $E_6=6$.

The computations use integer chains, integer inclusion matrices, and
finite-field characters. Section~\ref{app:certificates} of the supplement
lists the witnesses and the identities checked by each verification
program. The accompanying archive
\path{max7_arithmetic_revised.zip} includes the data, source code,
reproduction instructions, and checksums. It contains two forms of
network certificate: an integral filling with a constructive completion
proof for multiplier six, and a fully expanded weight array for
multiplier twenty-four.
\bibliographystyle{siamplain}
\bibliography{references}
\end{document}